\documentclass[12pt]{article} 

\usepackage[colorlinks=true]{hyperref}
\usepackage[margin=1.in]{geometry}
\usepackage{graphicx}
\usepackage{amsmath}
\usepackage{amssymb}
\usepackage{booktabs}
\usepackage{caption}
\usepackage{subcaption}
\usepackage{verbatim}
\usepackage{float}
\usepackage[english]{babel}
\usepackage{braket}
\usepackage{xy}

\title{A Classical-Quantum Convolutional Neural Network for Detecting Pneumonia from Chest Radiographs}

\author{Viraj Kulkarni \\ \small{Vishwakarma University}
        \and Sanjesh Pawale \\ \small{Vishwakarma University}
        \and Amit Kharat \\ \small{DeepTek Inc}}

\begin{document}
\hyphenpenalty=1000
\date{}

\maketitle

\begin{abstract}
\noindent While many quantum computing techniques for machine learning have been proposed, their performance on real-world datasets remains to be studied. In this paper, we explore how a variational quantum circuit could be integrated into a classical neural network for the problem of detecting pneumonia from chest radiographs. We substitute one layer of a classical convolutional neural network with a variational quantum circuit to create a hybrid neural network. We train both networks on an image dataset containing chest radiographs and benchmark their performance. To mitigate the influence of different sources of randomness in network training, we sample the results over multiple rounds. We show that the hybrid network outperforms the classical network on different performance measures, and that these improvements are statistically significant. Our work serves as an experimental demonstration of the potential of quantum computing to significantly improve neural network performance for real-world, non-trivial problems relevant to society and industry.
\end{abstract}

\section{Introduction}
Artificial neural networks have found widespread application in classification problems in the past decade. In a typical supervised classification setting, the training algorithm is provided a dataset of examples along with their associated labels. During the process of training, the algorithm extracts features from the examples and assigns different weights to each feature such that an overall loss function parameterized by these weights is minimized. The output of the training process is a function $f: \mathbb{R}^N \rightarrow \{1, 2, ..., k\}$, referred to as the trained model, that takes as input a set of features $x \in \mathbb{R}^N$ and predicts a class label $y \in \{1, 2, ..., k\}$ such that $y = f(x)$. The earliest neural networks used for image classification were fully connected multilayer perceptrons, where each neuron in a layer is connected to all neurons in the preceding and succeeding layers. These suffered from two limitations. First, by flattening a 2D image into a 1D vector before providing it as input, they did not take special advantage of the spatial relationship between the pixel data. Second, the full connectivity between neurons made them prone to overfitting.

Convolutional neural networks (CNNs) \cite{lecun1989backpropagation} exploit hierarchical patterns in the image data, where smaller and simpler patterns are assembled together to produce larger and more complex patterns. The input image is processed using a series of convolution operations implemented by convolutional layers, where each convolutional layer extracts patterns referred to as features from the output of the previous layer. Unlike fully connected networks, CNNs have sparser interactions between layers; and since the same convolution kernel is applied to all parts of the image, the network shares the same set of parameters for all locations in the image instead of learning a different set for each location. Due to this, CNNs can achieve good performance on machine learning tasks using several orders of magnitude fewer parameters than fully connected networks. Through the use of multiple layers stacked on top of one another, CNNs can learn complex representations at different levels of abstraction. In 2012, AlexNet - a CNN architecture - demonstrated a performance significantly superior to conventional machine learning approaches on the ImageNet dataset \cite{krizhevsky2012imagenet}. Since then, CNNs have found widespread applications for processing images, videos, and audio \cite{bhandare2016applications}. Various innovations have been proposed to the basic structure of the CNN over the last decade leading to networks such as VGG \cite{simonyan2014very}, YOLO \cite{redmon2016you}, Inception \cite{szegedy2015going}, U-Net \cite{ronneberger2015u}, ResNet \cite{he2016deep}, Mask RCNN \cite{he2017mask}, DenseNet \cite{huang2017densely}, etc.

The predictive power and expressibility of a neural network directly depends upon the number of its trainable parameters. Most networks in practice are heavily overparameterized; i.e. they contain more parameters than the number of training examples they are trained on \cite{soltanolkotabi2018theoretical}. Although architectural innovations do play an important role, most of the performance improvement on benchmark tasks in recent times can be attributed to two major factors: larger training datasets and larger network sizes. While both factors - the volume of training data and the sizes of neural networks - are both growing rapidly, the computational power available to us is no longer growing exponentially as it once used to. The informal Moore's law, which states that the number of transistors in a dense integrated circuit doubles every two years, is petering out \cite{leiserson2020there}. We are nearing the computational limits of deep learning, and further improvements in predictive performance  will require either markedly different computational methods or fundamental changes to the structure of neural networks \cite{thompson2020computational}.

Quantum computing has the potential to provide us with both. By exploiting quantum mechanical effects to execute multiple computational paths simultaneously, quantum computers could efficiently solve problems which are presently believed to be intractable for classical computers \cite{nielsen2002quantum}. On the other hand, by permitting transformations that are not possible to perform on classical computers, quantum computers may enable us to develop more expressive neural networks that learn better than classical ones \cite{servedio2004equivalences}. Thus, in other words, quantum computing may allow us to develop neural networks that not only train faster but are also more accurate and generalize better to unseen datasets \cite{abbas2021power}. Researchers have proposed a plethora of approaches that leverage quantum computing to enhance machine learning and neural network algorithms \cite{kulkarni2021quantum}\cite{biamonte2017quantum}\cite{dunjko2018machine}\cite{schuld2018supervised}. This excitement, however, is tempered by the fact that many of these techniques require quantum computational resources that are far beyond what today's quantum computers can offer.

A new area of research has emerged which specifically focuses on techniques that can be implemented on the noisy, intermediate-scale quantum (NISQ) computers that we anticipate will become available in the next few years \cite{perdomo2018opportunities}. Due to their smaller qubit counts and high error rates, NISQ hardware does not permit implementation of fully quantum neural networks. They do, however, support implementation of hybrid neural networks, where some layers in a classical neural network are replaced by variational quantum circuits (VQCs) - quantum circuits composed of unitary transformations that can be trained on a set of examples by tuning the circuit parameters \cite{havlivcek2019supervised}.

Researchers have recently implemented NISQ techniques for machine learning on experimental datasets such as CIFAR-10, MNIST, etc. \cite{ajayan2021edge}. The allure of NISQ techniques, however, lies in their promise to help us solve \textit{real-world, practical, industrially and commercially relevant} problems. In this paper, we present results of an experiment where we implement a classical-quantum hybrid classification network for detecting pneumonia from chest radiographs, a problem we believe meets this criterion. We demonstrate that substituting a classical layer with a quantum layer in a CNN while keeping constant the rest of the network architecture significantly improves the classification performance of the network. Neural networks are stochastic algorithms. Due to inherent randomness in the initialization of weights and biases, regularization techniques such as dropouts, optimization techniques such gradient descent, etc., different runs often yield widely varying results. While previous studies with hybrid neural networks compare results obtained in a single run, we repeat the experiment 30 times, thus making the findings statistically far more significant.

The rest of the paper is organized as follows. In section 2, we present the relevant background for quantum neural networks and usage of neural networks for automated interpretation of chest radiographs. Section 3 describes the datasets, experimental setup, and the results. In section 4, we discuss the results and provide a brief perspective on future work in this area.

\section{Background}

\subsection{Variational Quantum Circuits}

A variational quantum circuit consists of a series of quantum gates, each of which implements a unitary transformation on the input data. The parameters of such a circuit can be tuned to minimize a loss function based on how the circuit responds to examples in the training data; this makes the tuning process similar to training classical machine learning algorithms. Variational quantum circuits can thus be used for a wide variety of machine learning problems \cite{schuld2020circuit}\cite{farhi2018classification}. Furthermore, low-depth circuits are suitable for error correction \cite{li2017efficient} making them a natural choice for NISQ systems. In its general form, a variational quantum circuit $U(\theta)$, parameterized by a set of parameters $\theta$, is initialized to a \textit{zero} state. The output of the circuit can be represented by the observable $\hat{B}$. The loss function of this circuit can be defined as $f(\theta)=\bra{0}U^{\dag}(\theta)\hat{B}U(\theta)\ket{0}$. The circuit is trained by passing training examples through it while tuning the parameters $\theta$ to minimize the loss function $f(\theta)$. Like classical machine learning algorithms, this optimization is most commonly performed using a form of gradient descent. A theorical background of quantum circuit learning can be found in \cite{mcclean2016theory}\cite{mitarai2018quantum}. A feedforward neural network is composed of layers. Each layer can be thought of as a function that maps $n_{in}$ input vectors to $n_{out}$ output vectors. Such a layer in a classical neural network can be easily replaced by a variational quantum circuit forming a hybrid neural network. The quantum circuit itself could consist of multiple quantum layers with each layer representing one unitary transformation. Since all quantum transformations are unitary in nature, they maintain the dimensionality of the input. Hence, the dimension of the output for such a quantum layer is the same as the dimension of its input. 

At the classical-quantum input interface of the quantum layer, we need to embed the classical data vector into a quantum state. This is done by the embedding layer. Similarly, at the quantum-classical output interface of the quantum layer, we need to extract the classical output vector from the quantum output state. This operation is performed by the measurement layer. Thus, the quantum layer is a stack consisting of the embedding layer, the quantum operation, and the measurement layer. Typical embedding schemes encode each element of the classical data vector in one qubit. This creates a constraint on the quantum layer that the dimensionality of the input and output both need to be equal to the number of qubits constituting the quantum operation. A simple way around this limitation is to prepend a classical preprocessing layer before the quantum layer and to append a classical postprocessing layer after the quantum layer. Mari et al. \cite{mari2020transfer} refer to such a circuit as a \textit{dressed quantum circuit}. In this paper, we substitute a classical layer in a classical neural network by a dressed quantum circuit to create a classical-quantum-classical hybrid neural network.

\subsection{Automated Interpretation of Chest Radiographs}

Deep learning has, in recent years, evolved capabilities for performing detection \cite{lokwani2020automated}, segmentation \cite{hesamian2019deep}, and enhancement \cite{li2021review} for medical images. There have been several demonstrations of deep learning systems matching or exceeding performance of expert radiologists in predicting presence or absence of abnormalities from X-ray, computed tomography (CT), magnetic resonance (MR), and other types of radiology imaging scans \cite{qin2019using}\cite{ardila2019end}. The rapid progress in this field can be attributed to availability of two major factors which we lacked in the past: large, high-quality medical datasets and powerful computational hardware that can train large, complex neural networks. With growing research in the field, we expect datasets to grow larger over time and neural networks to get more complex. However, as discussed earlier, the computational power required to train these networks on ever-increasing volumes of data is approaching its theoretical limits, and we will need to look at different physical methods of computation to cater to these requirements in the future \cite{thompson2020computational}. Quantum computing is a leading candidate to do this.

In this paper, we choose the problem of detecting pneumonia from chest radiographs. Pneumonia is the leading cause of childhood mortality and kills approximately 2 million children every year \cite{rudan2008epidemiology}. The most common method of diagnosing pneumonia is by analyzing chest radiographs. However, owing to the worldwide shortage of qualified radiologists \cite{rimmer2017radiologist}, timely interpretation of radiographs is not always available, especially in low-resource geographical regions where the incidence of pneumonia is the highest. Systems providing automated detection of pneumonia from chest radiographs can facilitate quick and timely care for high-priority children and can be game-changer in such settings. Rajpurkar et al. \cite{rajpurkar2017chexnet} presented a 121-layer CNN that surpassed radiologist performance in detecting pneumonia from chest X-rays. The small-scale quantum computers and simulators available today, however, cannot run such large networks. Hence, we demonstrate the benefits of quantum computing on a smaller network that can be simulated. The following sections describe the dataset we used and the experimental setup.

\section{Experimental Setup}

\subsection{Data}

We used in our experiments a chest X-ray dataset published by Kermany et al. \cite{kermany2018identifying}. The dataset consists of 5,856 radiologist-labelled paediatric chest radiographs, out of which 1,583 are normal scans while 4,273 contain opacities suggestive of pneumonia. To enhance reproducibility of the experiments, we use a train-validation-test split offered by Kaggle \cite{breviglieri2020}, where the training subset contains 4,192 scans (1,082 normal and 3,110 pneumonia); the validation subset contains 1,040 scans (267 normal and 773 pneumonia); and the testing subset contains 624 scans (234 normal and 390 pneumonia. The original scans are 3-channel RGB images of varying resolution. We resize them to $128\times128$ and normalize them by dividing by 255, so all pixel values lie between 0 and 1.

\subsection{Classical Architecture}

\begin{figure}[ht]
\centering
\includegraphics[width=\linewidth]{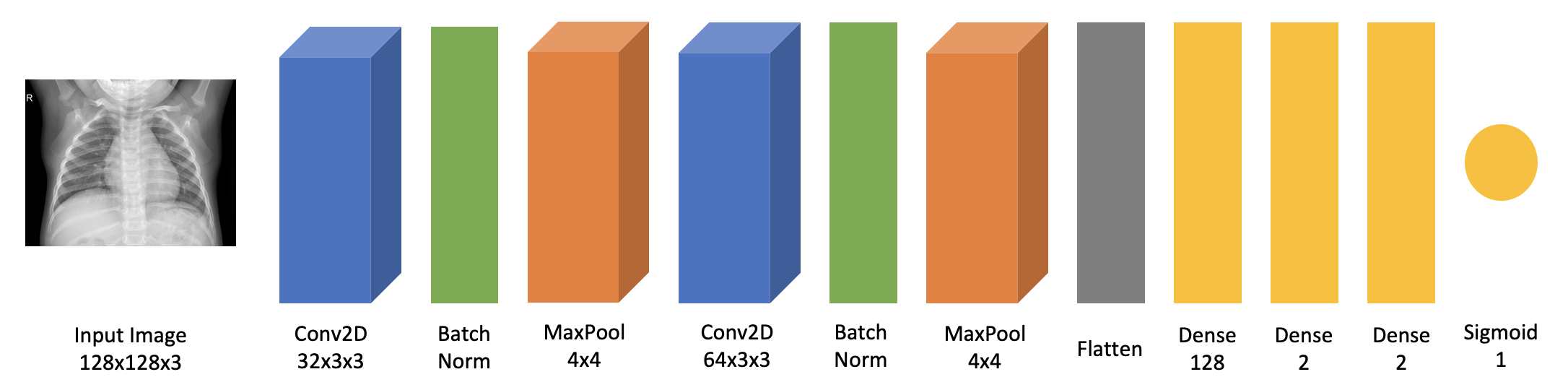}
\caption{Classical neural network architecture}
\label{fig:classical_architecture}
\end{figure}

The performance of a deep learning classification model is dominated by factors such as the volume of training data, accuracy of ground truth labels, size of the network, architecture of the network, etc. Our choice of network architure, however, was driven by the two different considerations: the architecture should be simple enough that replacing a classical layer by a quantum layer would have a meaningful impact; yet, it should also be powerful enough that it can perform the task of detecting pneumonia from chest X-rays with a better-than-random accuracy. In our pilot experiments, the simplest of CNNs consisting of just one convolutional block followed by a fully-connected layer did not perform well. We added layers to this network increasing its complexity to a point where it started learning from the training data (as seen by drop in the training and validation losses) and demonstrated a better-than-random accuracy on the test set. The architecture of this network can be seen in figure \ref{fig:classical_architecture}. The network had 11 layers with a total of 421,387 trainable parameters. We used a rectified linear unit (ReLU) as the activation function for all intermediate layers and sigmoid activation for the output layer. We used vanilla stochastic gradient descent as the optimizer keeping a constant learning rate of 0.01. We picked alternatives as simple as possible with the intention of mitigating the role these choices play on network performance. The network failed to learn with random initialization of its weights, so we initialized weights using the GlorotUniform initializer \cite{glorot2010understanding}.

\subsection{Hybrid Architecture}

\begin{figure}[ht]
\centering
\includegraphics[width=\linewidth]{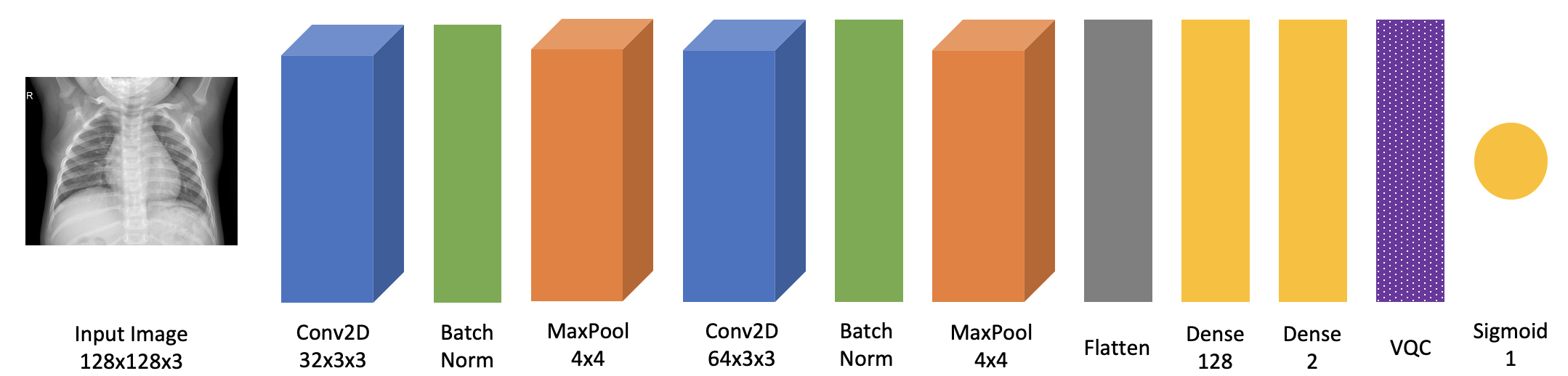}
\caption{Hybrid architecture replacing the penultimate layer with a quantum layer}
\label{fig:hybrid_architecture}
\end{figure}

We replaced the penultimate layer in the classical neural network by a quantum layer built out of a variational quantum circuit to create a hybrid neural network as shown in figure \ref{fig:hybrid_architecture}. All other hyperparameters were held constant between the two architectures. The penultimate layer, in the classical design, is a dense layer containing two neurons. Since the preceding layer has two neurons, and each neuron is associated with two weight parameters for the two inputs and one bias parameter, the penultimate layer consists of 6 trainable parameters. The capacity of a neural network to model a data distribution is closely related to the number of trainable parameters it has. Hence, to facilitate a fair comparison between performances of the classical and hybrid networks, we designed the quantum layer to also have 6 trainable parameters. Both classical and hybrid networks thus had the same number of trainable parameters.

\begin{figure}[ht]
\centering
\includegraphics[width=\linewidth]{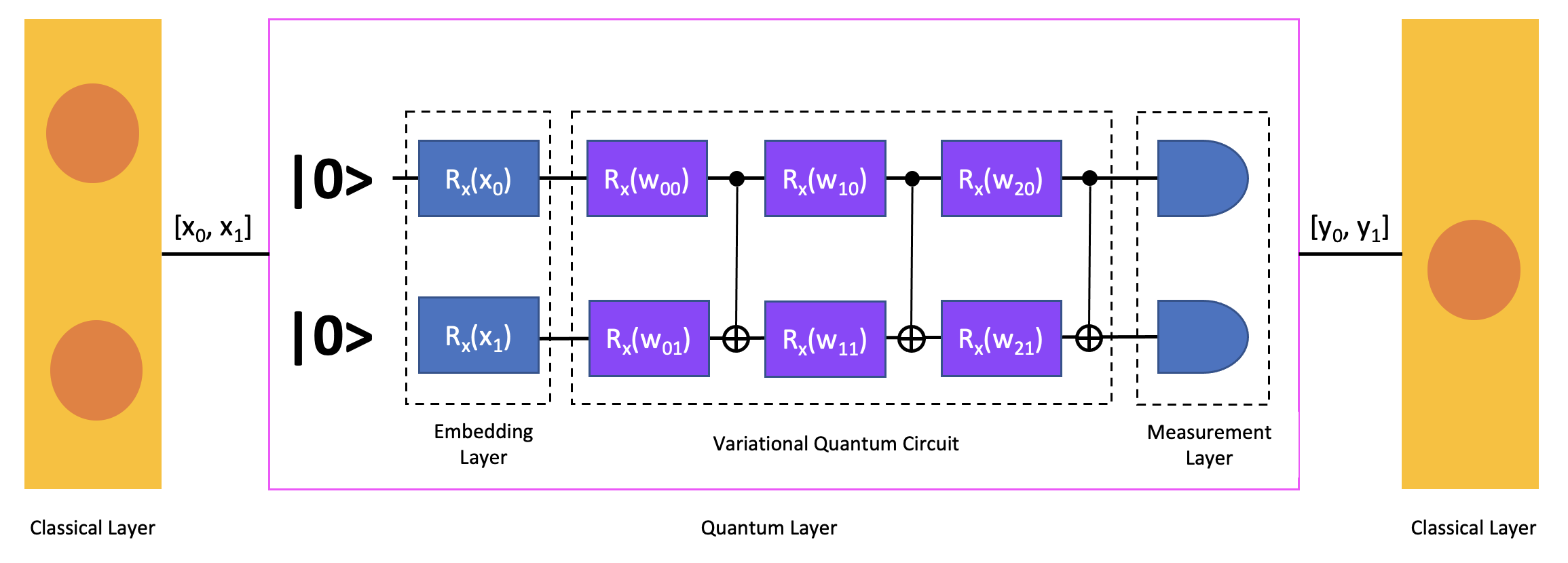}
\caption{Quantum layer consisting of the variational quantum circuit}
\label{fig:quantum_layer}
\end{figure}

The design of the variational quantum circuit is shown in figure \ref{fig:quantum_layer}. The features from the classical layer immediately preceding the quantum layer are encoded into two qubits using angle encoding. The value of the feature determines the angle of rotation about the X-axis as given by the $R_x$ gate: 

\begin{align}
R_x(\theta) = \begin{pmatrix}
\cos \frac{\theta}{2} & -i \sin \frac{\theta}{2} \\
-i \sin \frac{\theta}{2} & \cos \frac{\theta}{2}
\end{pmatrix}
\end{align}

The embedding layer thus converts the classical input vector $[x_0, x_1]$ into a two-qubit quantum state. This state is provided as input to the variational quantum circuit which consists of three parameterized, trainable rotation layers. Each layer implements a rotation about the X-axis on the system followed by an entangling CNOT operation: 

\begin{align}
CNOT = \begin{bmatrix} 1 & 0 & 0 & 0 \\ 0 & 1 & 0 & 0 \\ 0 & 0 & 0 & 1 \\ 0 & 0 & 1 & 0 \end{bmatrix}
\end{align}

Thus, there are a total of six trainable parameters in the circuit $\{w_{00}, ..., w_{21}\}$. Finally, the quantum state is converted back into a classical output vector $[y_0, y_1]$ through measurement by the Pauli-Z operation on the two qubits:

\begin{align}
Z \otimes Z = \begin{bmatrix} 1 & 0 & 0 & 0 \\ 0 & -1 & 0 & 0 \\ 0 & 0 & -1 & 0 \\ 0 & 0 & 0 & 1 \end{bmatrix}
\end{align}

\subsection{Implementation}

We wrote the software in Python 3 using Jupyter notebooks, used TensorFlow 2 \cite{tensorflow2015-whitepaper} and Keras \cite{chollet2015keras} for running the neural networks, and implemented the quantum layers in PennyLane \cite{bergholm2018pennylane} and IBM Qiskit \cite{Qiskit}. The execution of the code was done on a computer system with an Intel Core i7 processor and 32 GB RAM running Ubuntu 18.04 operating system. We simulated the quantum computations using the Pennylane quantum simulator developed by Xanadu Technologies.

Neural networks are stochastic systems that incorporate randomness at various stages. Activities such as initialization of weights and biases, regularization techniques like dropout, embedding methods, choosing starting points for gradient descent, etc. depend on a random number generator. These sources of randomness induce non-determinism and non-reproducibility in the results. Indeed, different runs of the same code using the same data often produce widely varying results \cite{d2020underspecification}. Some researchers bypass this limitation by fixing the seeds used for the random number generators. While this does improve reproducibility of the experiments, it does not aid meaningful comparison of two model architectures. To mitigate the influence of randomness on model performance, we compare performance of the architectures across multiple rounds. Sampling the results across multiple rounds also allow us to comment on the statistical significance of the results.

\subsection{Results}

Each round of our experiment consisted of training the model on the training dataset, recording the training loss and validation loss at the end of each epoch, and calculating accuracy and area under the receiver operating characteristics curve (AUROC) on the test dataset. We conducted 30 such rounds for the classical model and an equal number of rounds for the hybrid model. In both cases, we trained the model for 20 epochs.

\begin{figure}[ht]
\centering
\includegraphics[width=\linewidth]{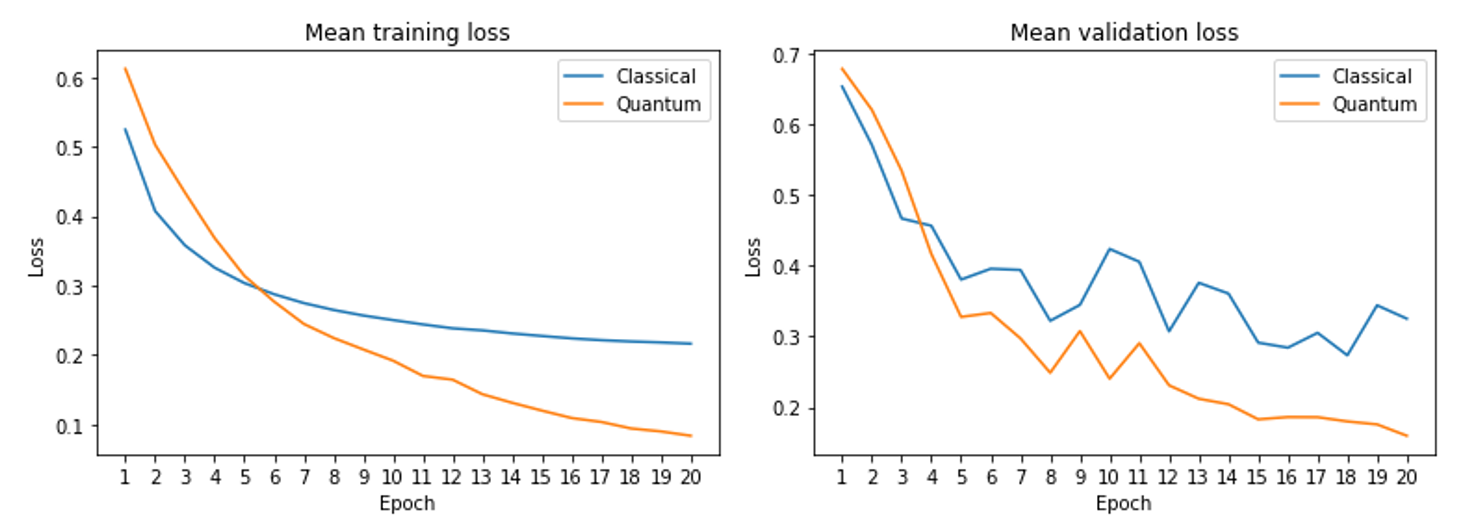}
\caption{Left: Training loss after each epoch; Right: Validation loss after each epoch}
\label{fig:plot_losses}
\end{figure}

The mean training and validation losses after each epoch averaged across all the rounds can be seen in figure \ref{fig:plot_losses}. The mean training loss at the end of the final epoch was 0.216 for the classical model and 0.083 for the hybrid model. The hybrid model thus learned to represent the training data better than the classical model as seen from the training losses. To test this hypothesis, we used the one-tailed Welsch's t-test and obtained a p-value of 0.012, thus showing that the result was statistically significant. We did the same comparison for validation losses after each epoch. The mean validation loss after the final epoch was 0.325 for the classical model and 0.160 for the hybrid model with a p-value of 0.002, thus demonstrating that the hybrid model was better than the classical model at generalizing to unseen data not used during training.

We further evaluated the models on the held-out test set. We obtained a mean AUROC score of 0.727 for the classical model and 0.836 for the hybrid model. Likewise, we obtained a mean accuracy of 0.706 for the classical model and 0.746 for the hybrid model. In both these cases, the improvements demonstrated by the hybrid model over the classical model were statistically significant. The detailed results can be seen in table \ref{tab:results}.

\begin{table}[ht]
\centering
\begin{tabular}{|l|l|l|l|l|}
\hline
Metric & Classical Architecture & Hybrid Architecture & p-value\\ 
\hline
Mean loss on training set & 0.216 & 0.083 & 0.012 \\
Mean loss on validation set & 0.325 & 0.160 & 0.002\\
AUROC on test set & 0.727 & 0.836 & 0.007\\
Accuracy on test set & 0.706 & 0.746 & 0.021\\
\hline
\end{tabular}
\caption{\label{tab:results}Performance comparison between classical and hybrid models}
\end{table}

\section{Discussion}

A plethora of techniques have been proposed to apply quantum computing to problems in machine learning. The performance of these techniques on real-world datasets, however, remains to be studied. Various quantum computing software libraries \cite{fingerhuth2018open} have emerged in recent years. These libraries enable researchers to rapidly develop software programs that implement these techniques and benchmark them against existing classical methods. In this paper, we explored how a variational quantum circuit could be incorporated within a classical neural network to create a hybrid neural network.  To compare the classical and hybrid networks, we chose the problem of detecting pneumonia from chest radiographs, since we consider this problem to be real-world, non-trivial, and relevant to society and industry. Our work serves as one of the first experimental demonstrations of the benefits quantum computing can bring to automated reading of medical images and, in general, another demonstration of quantum computing significantly improving neural network performance for real-world problems.

We simulated the quantum computations in software using the Pennylane quantum simulator. Several companies including IBM provide over-the-cloud access to their quantum computing hardware. Our research involved running 30 rounds of training and evaluation for the classical network and another 30 rounds for the hybrid network. While we would have liked to run our code on a physical quantum processor instead of a simulator, the scale of the experiment made this prohibitively expensive in terms of time. We leave this for future work.

The performance of a machine learning system in practice depends on a large number of considerations \cite{kulkarni2021jmir}. State-of-art neural network architectures for image classification contain millions of trainable parameters, and they are trained on hundreds of thousands of scans \cite{rajpurkar2017chexnet}. On the other hand, our network and the dataset it was trained on were both far smaller in size. We replaced only a single layer consisting of 6 trainable parameters with a variational quantum circuit to create the hybrid network. As quantum computing hardware becomes more accessible and the software ecosystem around it matures, researchers and developers would be able to incorporate larger quantum circuits within state-of-art classical neural networks and train these on large-scale datasets. We expect hybrid networks will begin outperforming the best-performing classical networks at that point. We see our work as a stepping stone leading towards that milestone.

\bibliographystyle{ieeetr}
\nocite{*}
\bibliography{bibliography}

\section*{Acknowledgements}
We acknowledge the use of TensorFlow, Keras, Pennylane, and IBM Quantum services for this work.

\end{document}